\documentclass{article}

\usepackage{arxiv}
\usepackage{multirow}
\usepackage[utf8]{inputenc} 
\usepackage[T1]{fontenc}    
\usepackage{hyperref}       
\usepackage{url}            
\usepackage{booktabs}       
\usepackage{amsfonts}       
\usepackage{nicefrac}       
\usepackage{microtype}      
\usepackage{lipsum}
\usepackage{graphicx}
\graphicspath{ {./images/} }
\usepackage{xcolor,colortbl}

\title{Augmenting Deep Neural Networks with Symbolic Knowledge: Towards Trustworthy and Interpretable AI for Education}

\author{
 Danial Hooshyar \\
  School of Digital Technologies\\
  Tallinn University\\
  Tallinn, 10120 \\
  \texttt{danial@tlu.ee} \\
   \And
 Roger Azevedo \\
  School of Modeling Simulation and Training\\
  University of Central Florida\\
  Orlando, FL 32826 \\
  \texttt{roger.azevedo@ucf.edu} \\
  \And
 Yeongwook Yang \\
  Division of Computer Engineering\\
  Hanshin University\\
  Osan, 18101 \\
  \texttt{yeongwook.yang@gmail.com} \\
}

\begin{document}
\maketitle
\begin{abstract}
Artificial neural networks (ANNs) have shown to be amongst the most important artificial intelligence (AI) techniques in educational applications, providing adaptive educational services. However, their educational potential is limited in practice due to three major challenges: i) difficulty in incorporating symbolic educational knowledge (e.g., causal relationships) in their development, ii) learning and reflecting biases, and iii) lack of interpretability. Given the high-risk nature of education, the integration of educational knowledge into ANNs becomes crucial for developing AI applications that adhere to essential educational restrictions, and provide interpretability over the predictions. This research argues that the neural-symbolic family of AI has the potential to address the named challenges. To this end, it adapts a neural-symbolic AI framework and accordingly develops an approach called NSAI, which injects and extracts educational knowledge into and from deep neural networks to model learners’ computational thinking. Our findings reveal that the NSAI approach has better generalizability compared to deep neural networks trained merely on training data, as well as training data augmented by SMOTE and autoencoder methods. More importantly, unlike the other models, the NSAI approach prioritizes robust representations that capture causal relationships between input features and output labels, ensuring safety in learning to avoid spurious correlations and control biases in training data. Furthermore, the NSAI approach enables the extraction of rules from the learned network, facilitating interpretation and reasoning about the path to predictions, as well as refining the initial educational knowledge. These findings imply that neural-symbolic AI can overcome the limitations of ANNs in education, enabling trustworthy and interpretable applications. 
\end{abstract}


\section{Introduction}
Recent advances in artificial intelligence (AI) have resulted in the proliferation of intelligent applications across various educational contexts, spanning both schools and broader educational systems \cite{vincent2020trustworthy}. At the core of these intelligent educational systems lies the \textit{learner model}, a crucial component that operates behind the user interface of learning environments \cite{conati20238}. By analysing data on learner-system interactions, the learner model generates a comprehensive representation of learners' knowledge and learning states, enabling the provision of adaptive and optimal learning experiences \cite{hooshyar2019systematic}. The process of constructing this learner model is known as \textit{learner modelling}, which relies on employing AI techniques, namely symbolic and sub-symbolic approaches \cite{abyaa2019learner}.  

Symbolic techniques offer the advantage of providing explanations and reasoning for the decisions reached within the learner model. However, they come with drawbacks such as high costs in terms of human involvement, necessitating the explicit embedding of real-world problems, and sensitivity towards data quality issues (\cite{garcez2022neural,lenat1985cyc}). On the other hand, sub-symbolic techniques require less human intervention, exhibit greater resilience towards noisy and incomplete data, and achieve higher predictive performance. Consequently, sub-symbolic methods like deep neural networks, which belong to the family of artificial neural networks (ANNs), have gained considerable popularity in various educational tasks, including learner modelling (e.g., \cite{bhanuse2021systematic,hernandez2019systematic,hooshyar2022gamedkt,piech2015deep}). ‌Despite their success and popularity, they face three primary challenges that limit their educational value. One of the main difficulties is incorporating educational constructs, restrictions, guidelines, causal relationships, and practitioners' knowledge (collectively referred to as \textit{educational knowledge}) into their development. Deep neural networks primarily operate with numerical data, requiring the translation of any other information into numerical values. This becomes problematic when modelling learners' behaviour and performance in digital learning due to the unavailability of representative training data or difficulty in collecting precise numerical data to capture adaptations of complex and dynamic learning patterns. Although incorporating educational knowledge can enhance training data, its integration into deep neural network applications in education is still limited.  

Second, like many other machine learning methods, deep neural networks are prone to learn misleading correlations during training, resulting in dependence on irrelevant or unnatural features which could lead to limiting their accuracy and generalizability \cite{torralba2011unbiased}. This could lead to significant failures when deploying the model in real-world applications (e.g., \cite{agrawal2018don,gretton2010consistent,srivastava2023addressing,zech2018variable}). In many cases, spurious correlations occur when a machine learning model relies on features that have a strong correlation with the target variable in the training data but are not causally related to it. To ensure fairness in AI applications in education, it is crucial to avoid algorithmic bias and prevent algorithms from being tuned to favour a majority group solely to achieve high-performance accuracy \cite{vincent2020trustworthy,unesco2019beijing}.

Finally, sub-symbolic methods, such as deep neural networks, struggle with explaining and reasoning their decision-making processes. Interpretability refers to the property of a learning model that explains its decisions in terms that humans can understand and provides insight into the reasoning process behind those decisions. In education, there is an urgent need to provide interpretability due to several reasons \cite{conati2018ai,meltzer2022european}. Firstly, a lack of interpretability hampers trust in deep neural network applications for educators and students. Both teachers and learners require transparency to comprehend the rationale behind recommendations, assessments, or feedback. Secondly, interpretability is crucial for effective learning. Students benefit from accurate and informative feedback. When AI systems cannot provide clear explanations for their recommendations or grading, students miss valuable learning opportunities. Thirdly, interpretability fosters critical thinking skills development. Education should nurture curiosity and encourage questioning and understanding of information and decisions. Uninterpretable deep neural networks hinder this process, limiting students' ability to question, analyse, or critique AI output. Lastly, the lack of interpretability can lead to bias and unfairness in educational AI systems. Without the ability to explain and understand the decision-making process, it becomes difficult to identify and rectify any biases that may exist in the data or the model itself. Recently, there have been several attempts to bring interpretability to deep neural network applications in education using common explainer algorithms like SHAP and LIME (e.g., \cite{hooshyar2022three,saarela2021explainable}). While helpful in generating an approximation of the decision, such techniques suffer from serious challenges (e.g., \cite{lakkaraju2020fool,slack2020fooling}). For instance, such explainer models can use a particular feature to predict without that feature appearing in the explanation of the prediction or often producing unrealistic scenarios. Most importantly, such models are unable to reason the path to their decisions.

To address these challenges, a prospective way is the recently emerging paradigm in the AI research community that is called \textit{neural-symbolic AI} or the \textit{3rd wave of AI}. By combining symbolic models with deep neural networks, neural-symbolic AI offers a potential solution to enhance learner modelling in intelligent educational systems (e.g., \cite{garcez2023neurosymbolic,hooshyar2021neural,sarker2021neuro}). This integration serves to leverage the interpretability and explanatory power of symbolic models while harnessing the scalability and predictive performance of deep neural networks. Consequently, neural-symbolic AI presents an opportunity to overcome the limitations of existing approaches and unlock new possibilities for more effective and efficient learner modelling \cite{hooshyar2021neural}. Despite their potential, their application in the field of education is extremely limited. Thus, this research aims to develop a neural-symbolic AI approach (hereafter called \textit{NSAI}) that has the potential to address the named challenges. The NSAI approach models learners’ computational thinking by incorporating both symbolic educational knowledge and training data in the construction of deep neural networks, and extracts hidden knowledge from the trained networks to explain and reason predictions. We compare the performance of the NSAI with multilayer perceptron (MLP) trained merely on training data, as well as training data augmented by SMOTE \cite{chawla2002smote} and autoencoder \cite{kingma2019introduction}. In order to achieve this aim, we set the following research questions:

\begin{itemize}
    \item How effectively can we ground educational knowledge in a deep neural network-based learner modelling approach in order to provide control over their behaviour?  
    
 \item How is the performance of the NSAI approach, in terms of generalizability, handling data biases, and interpretability of predictions, compared to deep neural networks? 

 \item What are the effects of data augmentation methods of SMOTE and autoencoders on the prediction power of deep neural network models?

\end{itemize}

\section{Related works}
\subsection{Neural networks in education}
\label{sec:headings}
Recently, there has been exponential growth in the use of artificial neural networks (ANN) in the educational context. Some examples of such applications are predicting student performance (e.g., \cite{bendangnuksung2018students,hooshyar2022gamedkt,piech2015deep,wang2017deep}), detecting undesirable student behaviour (e.g., \cite{fei2015temporal,teruel2018co,whitehill2017delving}), generating recommendations (e.g., \cite{abhinav2018lecore,bhanuse2021systematic,wong2018sequence}), evaluations (e.g., \cite{hooshyar2021predicting,taghipour2016neural,zhao2017memory}), and many more (for more details, see \cite{hernandez2019systematic}).

Regarding student performance prediction, knowledge tracing that revolves around the prediction of students' future performance based on their past activities is a significant challenge in education. Initial attempts by Piech et al. \cite{piech2015deep} introduced deep neural network (DNN) techniques, outperforming traditional machine learning methods, but their results faced scrutiny. Subsequent studies both supported (e.g., \cite{wang2017deep}) and challenged (e.g., \cite{mao2018deep}) Piech et al.'s work, comparing DNNs with traditional models and highlighting less significant differences. In a different context, DNN models were applied to analyse writing samples and clickstream data \cite{tang2016deep}, improve knowledge retention \cite{sharada2018modeling}, categorize learning capabilities (e.g., \cite{alam2018reduced}), and develop a sequential event prediction algorithm \cite{kim2018gritnet}. These studies demonstrated the effectiveness of DNNs in handling large student datasets and outperforming traditional approaches.  

Various works have addressed the detection of undesirable student behaviour in education, focusing on three subtasks: dropout prediction in MOOC platforms, evaluating social functions, and student engagement in learning. Regarding dropout prediction, studies applied DNN techniques, achieving superior performance compared to traditional machine learning methods. Dropout was defined and approached differently, including sequence labelling, binary classification, joint embedding, and personalized intervention models (e.g., \cite{teruel2018co,wang2017deep}). With regard to evaluating social function, Tato et al. \cite{tato2017convolutional} put forward a DNN-based approach for serious games that evaluates essential social ability (called sociomoral reasoning maturity) for adaptive social functioning. Concerning student engagement, DNN models utilizing various data sources such as facial video, gaze, voice, game trace logs, audio-visual information, and gaze and pose movements demonstrated high predictive accuracy (e.g., \cite{min2016predicting,sharma2016livelinet}). Additionally, DNN-based recommender systems were developed to recommend learning opportunities and personalized pathways based on students’ preferences and individual needs \cite{abhinav2018lecore,wong2018sequence}. Despite their success, the majority of the mentioned works are incapable of ensuring whether their developed DNN-based approach complies with educational knowledge and restrictions, do not properly take into account data bias-related issues (especially spurious correlations), underperform when it comes to datasets that are unrepresentative or small and are unable to provide explanation on their predictions in a way to be able to reason the path to the reached decisions. 

\subsection{Neural-symbolic AI}

Neural-symbolic AI has emerged as a noteworthy paradigm within the AI/Machine Learning research community \cite{garcez2022neural}. This approach is an active area of research that combines the principles of learning from experience with reasoning based on acquired knowledge. Basically, it integrates symbolic and connectionist (sub-symbolic) paradigms by representing knowledge symbolically and employing neural networks for learning and reasoning processes \cite{besold2017d}. Such integration allows for robust learning, logical reasoning, and interpretability. Recent related research has shown success in tackling existing challenges in the field of AI and machine learning. For instance, Tran and Garcez \cite{tran2016deep} developed a neural-symbolic approach that translates knowledge into network weights using confidence rules aimed at steering the learning process. Their findings reveal that their approach could improve the accuracy and generalizability of deep neural networks. In a similar attempt, Hu et al. \cite{hu2016harnessing} put forward a distillation framework for transferring knowledge (in the form of first-order logic rules) to neural networks. The difference between their work and Tran and Garcez's \cite{tran2016deep} was the way the symbolic knowledge was injected into the parameters of the neural networks using an iterative rule distillation process. Their results show how such knowledge injection could improve the generalizability of the network compared to baseline models. Serafini and Garcez \cite{serafini2016learning} propose a novel framework for learning and reasoning that incorporates symbolic knowledge by transforming the loss function of the deep neural networks. Their findings show that their approach could be successfully used for various data prediction and knowledge completion tasks. Li and Srikumar \cite{li2019augmenting} propose constraint-based architectures for embedding knowledge in neural network architectures. They achieved this by translating first-order logic into differentiable components of the networks, without the need for additional learnable parameters. Results of their experiment indicate their success in equipping networks with domain knowledge to relax challenges like limited training data. Despite being extensively studied in various fields, the use of neural-symbolic AI in education is still limited.

Recently, Hooshyar and Yang \cite{hooshyar2021neural} proposed a framework for bringing together the principles of neural-symbolic AI and applying them to the domain of education. This framework provides a solid basis for creating AI solutions that are interpretable and designed specifically for educational environments. Furthermore, Shakya et al. \cite{shakya2021student} developed a neural-symbolic approach that combines the semantics of symbolic models, such as Markov Logic, with deep neural networks like LSTM. By leveraging the graph structure encoded by Markov Logic, the model learns symmetries and efficiently trains the LSTM using importance sampling. The evaluation of KDD EDM challenge datasets demonstrates the superiority of the neural-symbolic model over HMMs and pure LSTM methods, achieving high prediction accuracy by focusing on a smaller fraction of the training data. This research aims to build upon existing works by proposing a neural-symbolic AI approach called NSAI, which not only injects educational knowledge but also transforms the model architecture \cite{muralidhar2018incorporating}. NSAI allows explicit integration of propositional educational knowledge during training, providing control over the model's behaviour. Additionally, the NSAI approach addresses data bias issues, specifically spurious correlations, ensuring algorithmic fairness and compensating for a lack of training data and data inconsistencies. Finally, our approach enables the extraction of knowledge from trained deep neural networks, enhancing their applications with interpretability.

\section{Neural-symbolic AI for modelling learners’ computational thinking}
\subsection{Context on the AutoThinking game for computational thinking}
AutoThinking is an educational game designed to improve learners' computational thinking abilities. Instead of using traditional programming languages, the game utilizes icons to represent programming concepts, eliminating the chance of syntax errors. AutoThinking stands out by incorporating adaptivity in both gameplay and the learning process, making it the first of its kind to foster computational thinking skills. The game focuses on four essential CT skills: breaking down problems into smaller steps (algorithmic thinking), constructing algorithms through pattern recognition and generalization, identifying and fixing errors (debugging), and simulating solutions. It also introduces three fundamental programming concepts: sequencing actions, making decisions based on conditions, and repeating actions in loops \cite{hooshyar2019autothinking}.

\begin{figure}
  \centering
  \begin{tabular}{@{}c@{}}
    \includegraphics[width=\textwidth]{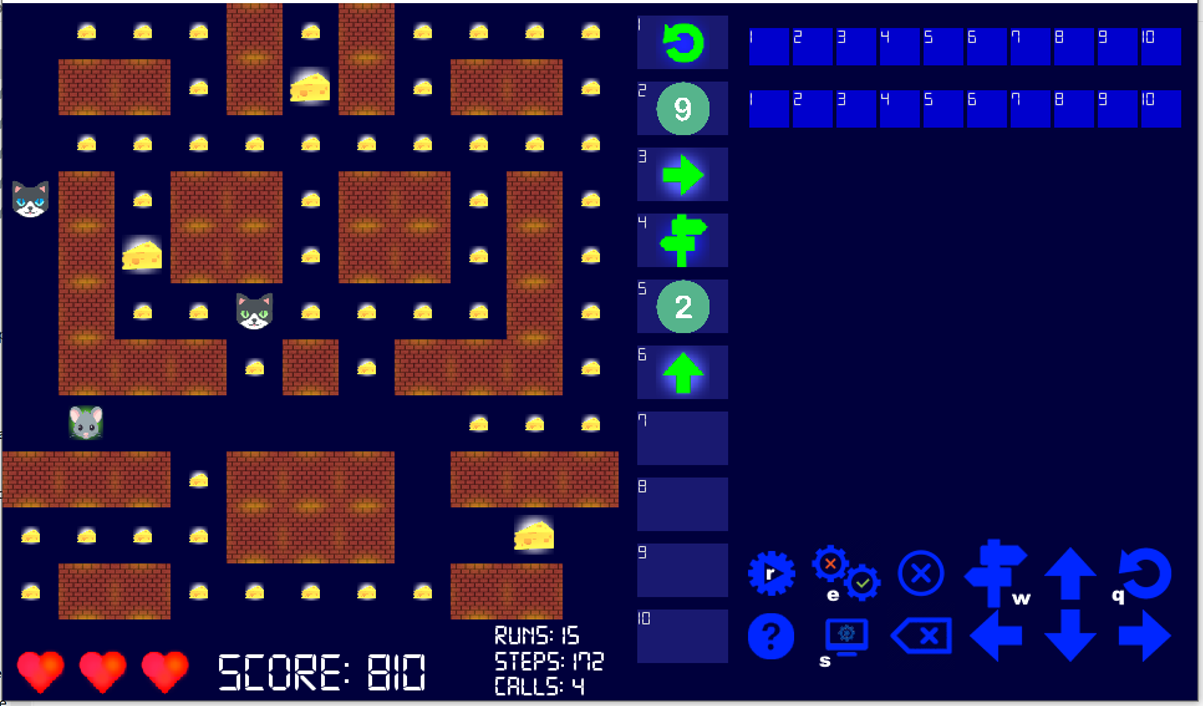} \\[\abovecaptionskip]
    \normalsize (a)
  \end{tabular}
  \begin{tabular}{@{}c@{}}
    \includegraphics[width=\textwidth]{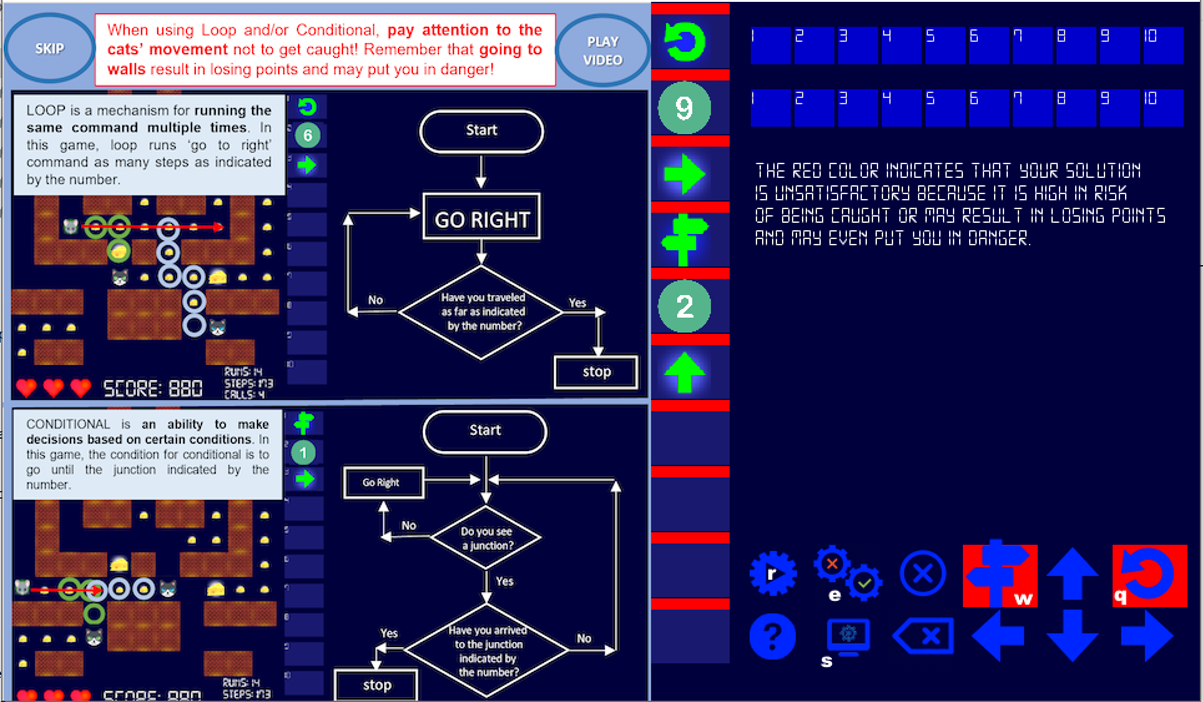} \\[\abovecaptionskip]
    \normalsize (b)
  \end{tabular}

  \caption{(a) a solution developed by a player, and (b) feedback and a hint generated for the solution (taken from Hooshyar et al., \cite{hooshyar2021gaming}).}\label{fig:myfig1}
\end{figure}

In the game, players take on the role of a mouse and navigate through different levels. Their main objectives are to collect cheese pieces, earn points, and avoid cats in a maze. They can come up with up to 20 different strategies to collect all 76 cheese pieces. Higher scores are awarded for solutions that incorporate critical thinking concepts and skills, as well as for navigating through non-empty tiles. Players have the flexibility to develop various solutions, including using functions to save and apply patterns in different situations, and the game provides adaptive feedback and hints. Examples of a learner-developed solution, as well as feedback and hints generated by the game are illustrated in Figure~\ref{fig:myfig1}a and b, respectively. 

\subsection{The proposed NSAI approach}
To showcase the feasibility of applying NSAI frameworks for injecting educational knowledge in addition to training data into the development of (deep) neural networks, we adapt the KBANN framework developed by \cite{towell1994knowledge}. Briefly, the approach allows incorporating symbolic knowledge into the architecture of neural networks so as to augment the training data, control the training flow of the network, and provide context to map the learned representation into symbolic knowledge which interprets and reasons the decision-making of the network. Table~\ref{tab:table1} presents the algorithm of the framework. 

The NSAI approach begins with loading the training data and the educational knowledge in the form of rules (see Datasets and educational knowledge section). It then implements Towell's rewriting algorithm and if there is more than one rule to consequent, then rewrite it as two rules \cite{towell1994knowledge}. It then establishes a mapping between the set of rules and the neural network, in a way to create layers, weights and biases for the neural network. Finally, it applies Backpropagation using training examples and uses weights and biases of the learned network to extract rules, explaining the predictions. 
\begin{table}[!htb]
\centering
\small
\caption{Overall algorithm of the KBANN framework used in the proposed NSAI approach}
\begin{tabular}{|c|p{.8\linewidth}|}
\hline
Step    & \multicolumn{1}{c|}{Description}\\
\hline
$1$      & Rewrite rules (the symbolic knowledge in propositional logic form) to eliminate disjuncts \\ \hline

$2$       & Map rule structure into a neural network\\ \hline

$3$       & Add important features not specified in mapping\\ \hline
$4$           & Add hidden units to the neural network \\ \hline
$5$           & Label units in the KBANN according to their level \\ \hline
$6$           & Add links not specified by translation between all units in topologically contiguous levels \\ \hline
$7$           & Perturb the network by adding near-zero random numbers to all link weights and biases \\ \hline
$8$           & Assign high-weight values to the links created from the domain knowledge rules \\ \hline
$9$           & Apply Backpropagation to refine the network to fit the training data \\ \hline
$10$           & Use weights and biases of the learned network to extract rules, explaining the predictions \\
\hline
\end{tabular}
\label{tab:table1}
\end{table}

\begin{figure}
  \centering
  \begin{tabular}{@{}c@{}}
    \includegraphics[width=\textwidth]{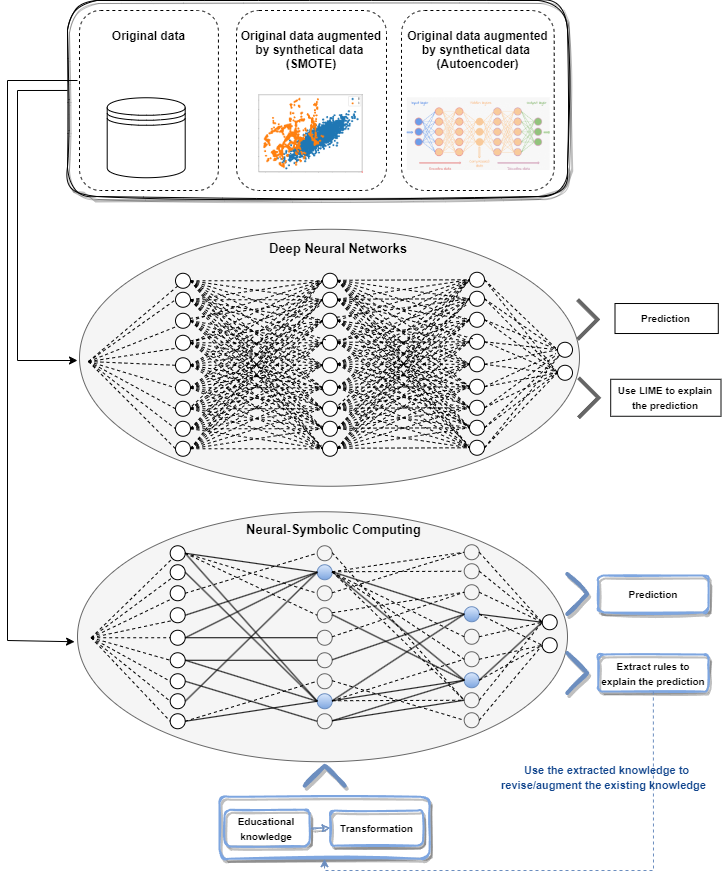} \\[\abovecaptionskip]
  \end{tabular}
  \vspace{\floatsep}
  \caption{The overall architecture of the NSAI approach.}\label{fig:myfig2}
\end{figure}

Figure~\ref{fig:myfig2} illustrates the overall architecture of the NSAI approach. As shown in the figure, we first generate synthetical data to augment the training data using two methods of SMOTE Upsampling and autoencoders (see section Experiment setting and evaluation). Thereafter, we train and evaluate the performance of the deep neural network using the three different learning sources of original data (called Deep NN), original augmented by SMOTE (called Deep NN-SMOTE), and original augmented by autoencoder (called Deep NN-Autoencoder). Additionally, we implement the LIME method \cite{ribeiro2016model} to provide local and global explanations for the predictions. On the other hand, we developed the NSAI approach that uses educational knowledge in addition to the original data. Upon training and evaluation, it also extracts rules from the networks to explain the predictions and reason the path to the decisions.  

\textbf{\textit{Hypothetical example:}} Let’s assume we have the below domain knowledge and training examples, see Table~\ref{tab:table2}, related to self-regulated learning (SRL). For more information on the multilevel decomposition of SRL strategies involved during multimedia learning, see Azevedo and Dever \cite{azevedo201410}, Greene and Azevedo \cite{greene2009macro}, Pekrun \cite{pekrun2006control}, and Winne and Azevedo \cite{Winne2022}. 

\textbf{\textit{Symbolic domain knowledge: }}

\hspace{1em}\textit{Final performance :- Cognition, Metacognition, Emotion, Motivation.} \vspace{-1ex}

\hspace{1em}\textit{Cognition :- Planning, Search for information, Making inferences.} \vspace{-1ex}

\hspace{1em}\textit{Metacognition :- Goal setting, Information structuring, Judgement of learning.} \vspace{-1ex}

\hspace{1em}\textit{Information structuring :- Monitoring progress towards goals. } \vspace{-1ex}

\hspace{1em}\textit{Emotion:- Help seeking, Self-reported emotion rating.  } \vspace{-1ex}

\hspace{1em}\textit{Motivation :- Time watching learning materials, Forum chat. }

\begin{table}[!htb]
\centering
\small
\caption{Training examples}
\begin{tabular}{|l|l|l|l|l|l|l|l|l|l|l|} 
\hline
\multirow{2}{*}{Features} & \multicolumn{10}{c|}{Label (Final Performance)}                 \\ 
\cline{2-11}
& \multicolumn{4}{c|}{Low} & \multicolumn{6}{c|}{High}  \\ 
\hline
Goal Setting                        &                       &   &   & x &   &   & x & x & x &       \\ \hline
Prior Knowledge Activation          &                       &   & x &   & x &   &   &   &   &       \\ \hline
Planning                            & x                     &   &   & x & x & x & x & x & x & x     \\ \hline
Judgement of Learning               & x                     & x &   &   & x & x & x & x & x & x     \\ \hline
Time Management                     &                       & x &   &   &   &   & x & x & x &       \\ \hline
Monitoring (Progress Towards Goals) &                       & x & x &   &   &   &   & x &   & x     \\ \hline
Search for Information              & x                     &   &   &   &   & x & x &   &   & x     \\ \hline
Help Seeking                        &                       & x & x &   & x & x &   &   &   & x     \\ \hline
Frustration                         & x                     & x &   &   &   & x & x &   & x &       \\ \hline
Time Watching Learning Materials    &                       & x & x &   &   &   &   & x & x &       \\ \hline
Self-reported Emotion Rating        &                       & x &   &   &   &   & x &   &   &       \\ \hline
Concentration                       & \multicolumn{1}{c|}{x} &   & x &   &   & x & x &   &   & x     \\ \hline
Forum Chat                          &                       &   &   &   &   & x & x &   &   &       \\ \hline
Automatic Logout                    &                       & x & x & x &   &   &   &   &   &       \\ \hline
Making Inferences                   &                       &   &   & x &   &   & x &   & x &       \\
\hline
\end{tabular}
\label{tab:table2}
\end{table}

The first step is to employ the domain knowledge (which could be in the form of rules, logical relationships, etc.) to initialize the network architecture. This involves determining the number and types of layers, the number of neurons in each layer, and the connections between them. The architecture is constructed to reflect the problem-specific characteristics and constraints captured by the domain knowledge. Once the architecture is defined, the domain knowledge is encoded into the network by incorporating the expert rules or constraints into the structure or parameters of the neural network. For example, certain connections may be fixed or constrained based on specific rules or relationships from the domain knowledge. After initializing the network with domain knowledge, the training process begins. Initially, the network's parameters (weights and biases) are typically randomly assigned or initialized. The training data, consisting of input-output pairs, is presented to the network. Thereafter, forward propagation and error calculation take place, and then backpropagation to adjust the network's parameters to reduce the overall error. This process repeats for multiple iterations or epochs, allowing the network to gradually improve its performance by learning from the training data and adjusting its parameters based on the error feedback. The training process continues until a termination criterion is met.

\begin{figure}
  \centering
  \begin{tabular}{@{}c@{}}
    \includegraphics[width=\textwidth]{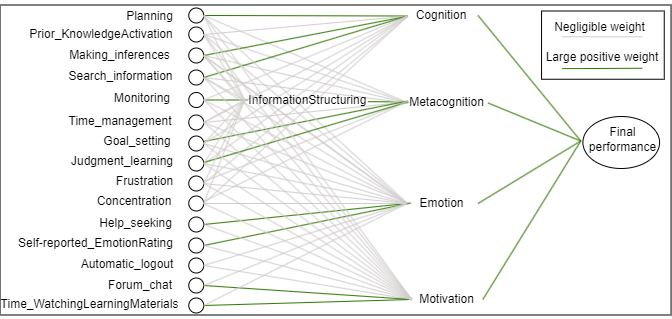} \\[\abovecaptionskip]
    \normalsize (a)
  \end{tabular}
  \begin{tabular}{@{}c@{}}
    \includegraphics[width=\textwidth]{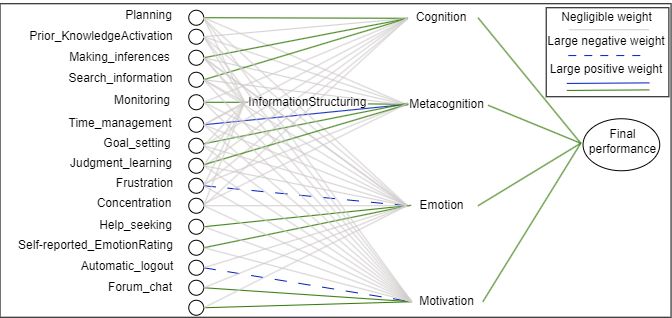} \\[\abovecaptionskip]
    \normalsize (b)
  \end{tabular}
  \caption{(a) initializing the network using the domain knowledge, and (b) the adjusted network after training.}\label{fig:myfig3}
\end{figure}

In our example, as shown in Figure~\ref{fig:myfig3}a, the network is initialized using the domain knowledge. This is shown using solid fixed connections between the SRL components and input features. For instance, the green line between \textit{Cognition} and \textit{Search for Information} indicates that learners with rather good information-searching skills during digital learning tend to have better cognitive skills. Similarly, for each rule in the domain knowledge, there is a fixed connection with rather large weights between the features and the latent variable. Additionally, there are connections with negligible weights. These are connections that have very small weights, close to zero, after the initialization or during training. These connections may have minimal impact on the network's output or learning process. After the initialization, the network is adjusted using the training examples. As Figure~\ref{fig:myfig3}b shows, the initial network has learned from the training examples to create a fixed connection with a high weight between \textit{Metacognition} and \textit{Time Management}. Moreover, it learned to create a fixed connection with large negative weights between \textit{Motivation} and \textit{Automatic Logout}, and \textit{Emotion} and \textit{Frustration}. This indicates that learners tend to have better affective states if they have lower automatic logout and frustration during digital learning (for more details, see \cite{towell1994knowledge}).  

\section{Results and analysis}
\subsection{Datasets and educational knowledge}
To model learners’ computational thinking during gameplay, we used data from 427 players in five different countries (Estonia, France, South Korea, Taiwan, and South Africa) who were playing the third level of the AutoThinking game. Studies conducted by El Mawas et al. \cite{el2020investigating} and Hooshyar et al. \cite{hooshyar2021gaming} show some study examples of how the data were collected.  During the gameplay, the system records different types of learner interactions. These include: 1) tracking the mouse and non-player characters' positions, 2) task identifiers, 3) the small and big cheese collection, 4) the usage of loops, conditionals, arrow, and function, 5) debugging and simulation activity, 6) the frequency of seeking help, 7) the number of feedback and hints, 8) the frequency of colliding with walls, 9) the estimation of the learner's CT knowledge, and 10) the evaluation of solution quality inferred from the Bayesian network decision-making algorithm employed in the game. More information regarding the decision-making process is given in the work by Hooshyar, Lim, et al. \cite{hooshyar2019autothinking}. For this research, we selected those features that are directly/indirectly related to players’ performance. Table~\ref{tab:table3} and Figure~\ref{fig:myfig4} provide a brief summary and distribution of the datasets, respectively. 

\begin{figure}
  \centering
  \begin{tabular}{@{}c@{}}
    \includegraphics[width=\textwidth]{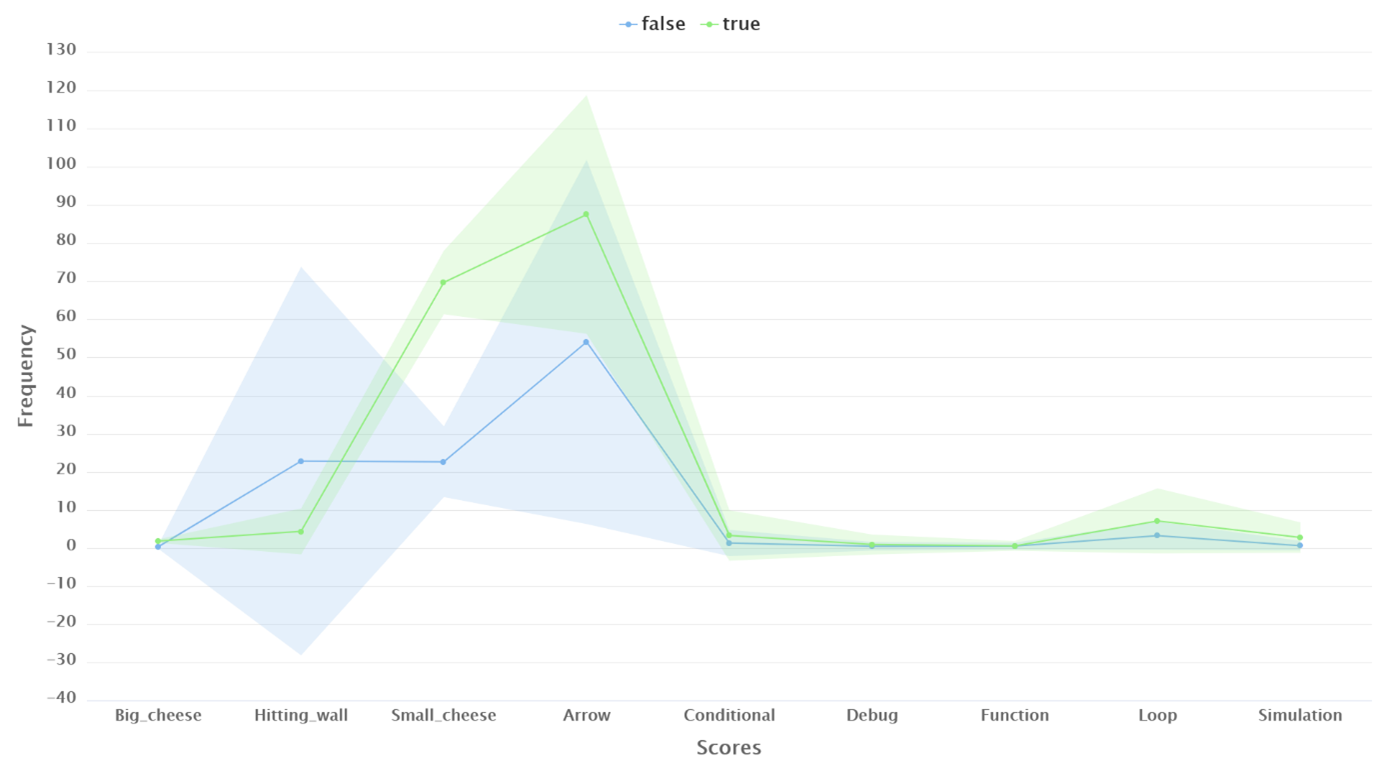} \\[\abovecaptionskip]
    \normalsize (a)
  \end{tabular}
  \vspace{\floatsep}
  \begin{tabular}{@{}c@{}}
    \includegraphics[width=\textwidth]{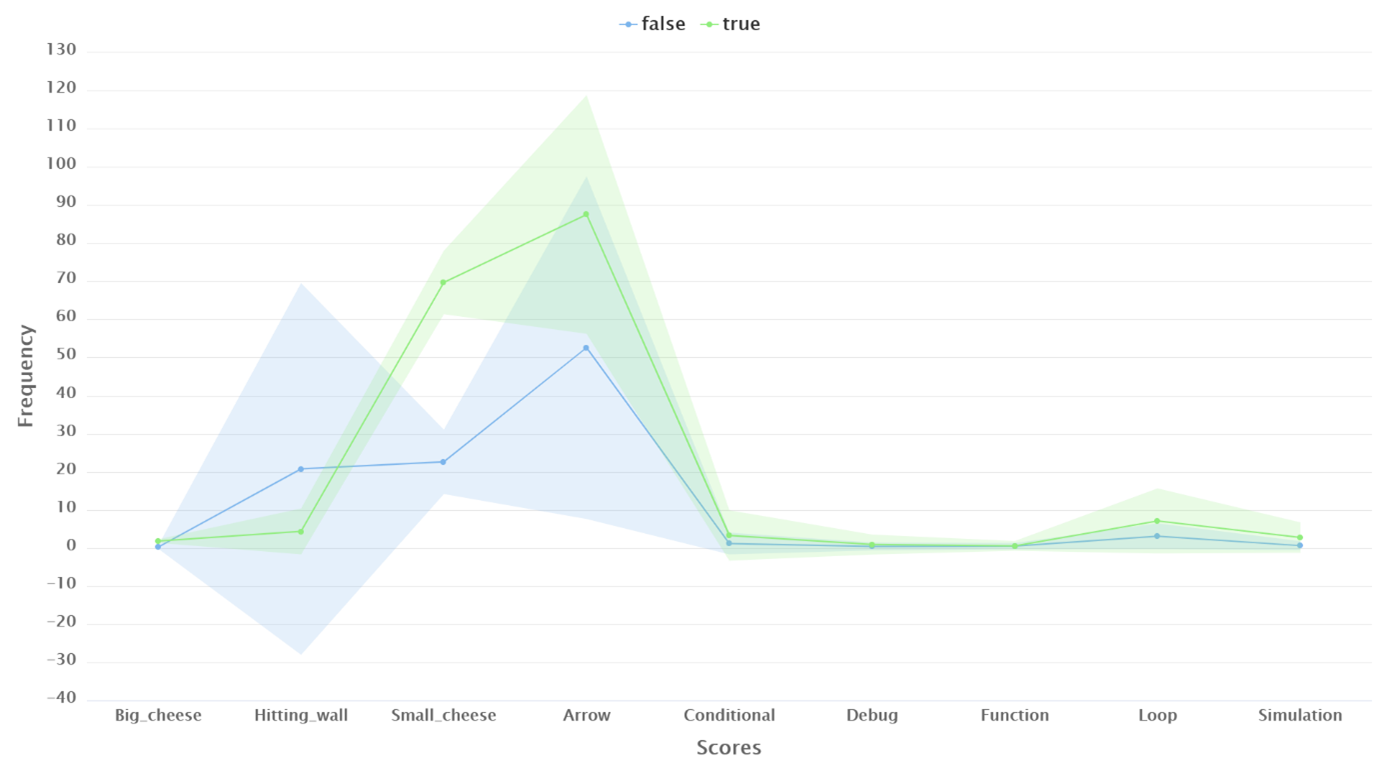} \\[\abovecaptionskip]
    \normalsize (b)
  \end{tabular}
\end{figure}
\clearpage
\begin{figure}
  \begin{tabular}{@{}c@{}}
    \includegraphics[width=\textwidth]{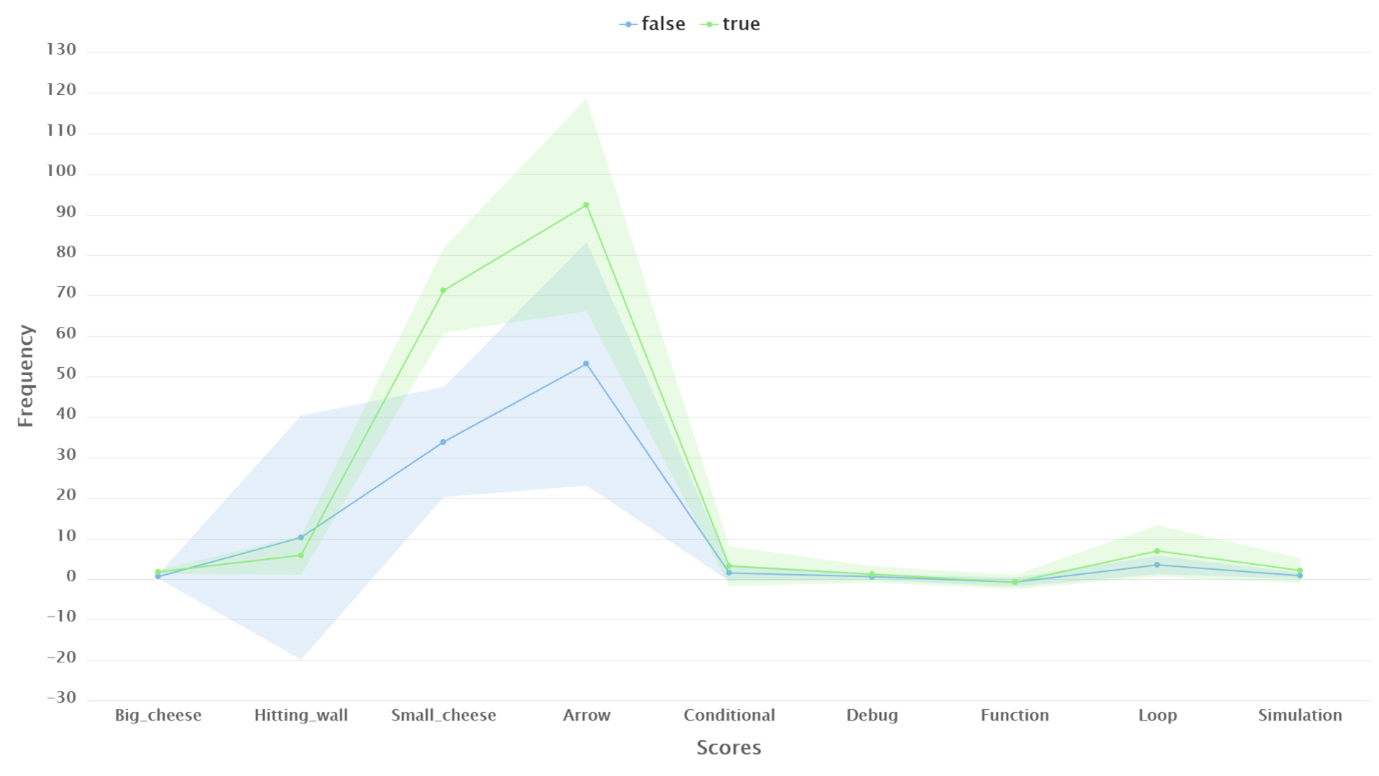} \\[\abovecaptionskip]
    \normalsize (c)
  \end{tabular}
    \begin{tabular}{@{}c@{}}
    \includegraphics[width=\textwidth]{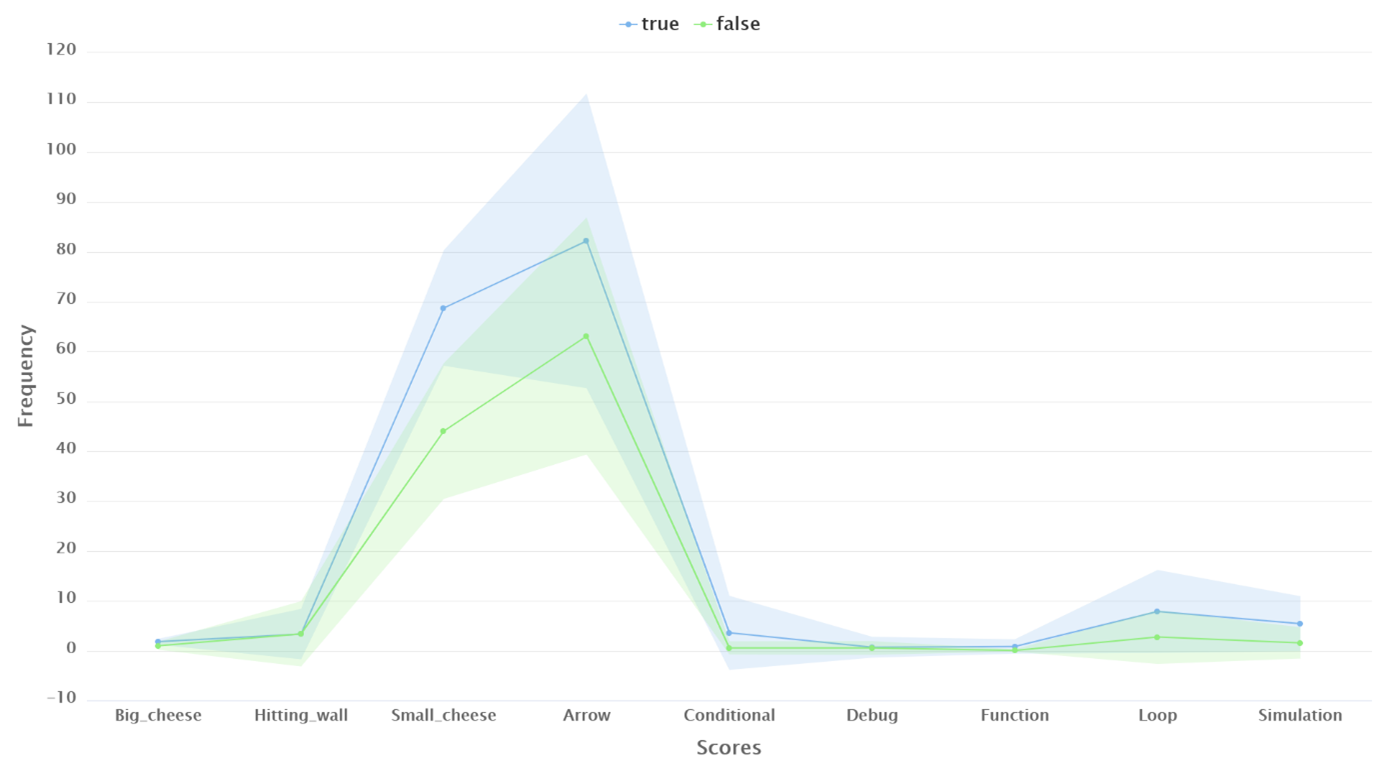} \\[\abovecaptionskip]
    \normalsize (d)
  \end{tabular}
  \caption{Deviation chart showing the distribution of: a) the original training, b) training augmented with SMOTE, c) training augmented with Autoencoder, and d) test data.}
        \label{fig:myfig4}
\end{figure}
\clearpage

\begin{table}[!htb]
\centering
\small
\caption{Distribution of the dataset before splitting}
\begin{tabular}{|l|l|l|l|l|} 
\hline
Features     & Min & Max                      & Average & Deviation             \\ 
\hline
Arrow        & 15  & 180                      & 82.05   & 34.65                 \\ \hline
Big cheese   & 0   & 4                        & 1.6     & 0.7                   \\ \hline
Small cheese & 0   & 74                       & 63.38   & 17.72                 \\ \hline
Function     & 0   & 4                        & 0.6     & 1.2                   \\ \hline
Debug        & 0   & 17                       & 0.8     & 2.3                   \\ \hline
Simulation   & 0   & 19                       & 2.92    & 4.24                  \\ \hline
Loop         & 0   & 50                       & 6.66    & 8.12                  \\ \hline
Conditional  & 0   & 46                       & 3       & 6.4                   \\ \hline
Hitting wall & 0   & 180                      & 6.19    & 18.57                 \\ 
\hline
Final score  & \multicolumn{2}{c|}{True = 364} & \multicolumn{2}{c|}{False = 63}  \\
\hline
\end{tabular}
\label{tab:table3}
\end{table}

To introduce domain knowledge into the NSAI approach, we adapted CT skills and concepts used by Hooshyar \cite{hooshyar2022effects}. This includes CT skills of problem identification, building algorithms, debugging, simulation; and concepts of sequences, loops, and conditional. To present the knowledge into the neural networks, we used propositional logic (i.e., a set of propositional non-recursive Horn clauses), shown in Table~\ref{tab:table4}. Briefly, the first propositional logic indicates that the final score or performance in the game depends on the mastery of CT skills and concepts (for more details, see \cite{hooshyar2022effects}). Additionally, it states that the CT concepts are associated with features of conditional and loop in the training dataset, whereas the CT skills are dependent on debug, simulation and function features. In other words, learners’ overall performance in CT is good if they are good at CT concepts and skills. 
\begin{table}[!htb]
\centering
\small
\caption{Educational knowledge represented in symbolic form of rules}
\begin{tabular}{|p{0.7\linewidth}|}
\hline
\texttt{Final\_score:- CT\_concepts, CT\_skills.} \\ \hline
\texttt{CT\_concepts:- Conditional, Loop.} \\ \hline
\texttt{CT\_skills:- Debug, Simulation, Function.} \\
\hline
\end{tabular}
\label{tab:table4}
\end{table}
\subsection{Data bias-related issues}
In the context of a training dataset, data bias refers to the presence of systematic and non-random errors or distortions in the data that can impact the performance and fairness of machine learning models. It occurs when certain subsets or categories within the dataset are overrepresented or underrepresented, leading to a skewed understanding of the underlying patterns and relationships. When a dataset is biased, it fails to accurately reflect the intended application of a model, leading to distorted results, reduced accuracy, and analytical mistakes.

Data bias in a training dataset can arise due to various factors, such as the sampling process, data collection methods, presence of confounding variables, etc (European Union Agency for Fundamental Rights, \cite{europaBiasAlgorithms}). These issues can impact the performance and fairness of machine learning models. Recently, there have been some research studying data biases and their effect on the performance of machine learning models. For instance, Blodgett et al. \cite{blodgett2020language}’s work mainly revolves around ensuring that models do not rely on sensitive features such as race and gender, and Johnson and Khoshgoftaar \cite{johnson2019survey} focus on addressing class imbalance and skewed distributions. However, when it comes to education, despite the high-risk nature of the education domain and the widespread use of ANNs in education, there is not much research taking into account spurious correlations biases and their effect on the performance of machine learning (especially ANNs). Spurious correlations pose a major challenge when deploying machine learning models because they can cause the models to depend on irrelevant or unnatural features, leading to significant failures when deploying the model in real-world applications (e.g., \cite{agrawal2018don,gretton2010consistent,zech2018variable}).

‌In many cases, spurious correlations occur when a machine learning model relies on features that have a strong correlation with the target variable in the training data but are not causally related to it. For instance, in sentiment classification, a bias in the training data can arise where positive examples tend to be longer than negative examples. In such cases, the model may erroneously consider length as a crucial feature for classification, even though it is a spurious feature that does not actually provide sentiment information \cite{liusie2022analyzing}. In case of the AutoThinking game, an example could be the existence of a strong correlation between the input variable of Small\_cheese and the label (Final\_score). Although the inclusion of the Small\_cheese feature in the model could offer some useful information, it is important to note that there is no causal relationship between this feature and the class label. This is because many players employ random strategies in the game, where a solution can be successful in collecting both small and big cheeses without necessarily utilizing main CT skills and concepts. For instance, a solution that solely uses arrows to navigate the game without the NPC catching the mouse can still collect many Small\_cheese and accordingly achieve high scores. In educational games, a high score can be obtained through either a random strategy or an appropriate strategy, such as parallel thinking \cite{hooshyar2023modeling}. Consequently, Table~\ref{tab:table3} and ~\ref{tab:table5} indicate the presence of class imbalance and potential data biases of spurious correlation in the training dataset, respectively.
\begin{table}[!htb]
\small
\centering
\caption{Correlation between features and class label in training and test data}
\begin{tabular}{|l|l|l|l|l|} 
\hline
\multirow{2}{*}{Features} & \multicolumn{4}{c!{\color{black}\vrule}}{Final score}                                     \\ 
\cline{2-5}
                          & Train data & SMOTE augmented train data & Autoencoder augmented
  train data & Test data  \\ 
\hline
Arrow                     & 0.322     & 0.412                      & 0.51                               & 0.255      \\ 
\hline
Big\_cheese               & 0.728      & 0.858                      & 0.68                               & 0.445      \\ 
\hline
Conditional               & 0.107      & 0.203                      & 0.154                              & 0.175      \\ 
\hline
Debug                     & 0.066      & 0.124                      & 0.135                              & 0.027      \\ 
\hline
Function                  & 0.011      & 0.027                      & 0.005                              & 0.233      \\ 
\hline
Hitting\_wall             & -0.31      & -0.23                      & -0.122                             & -0.003     \\ 
\hline
Loop                      & 0.164      & 0.295                      & 0.239                              & 0.25       \\ 
\hline
Simulation                & 0.194      & 0.343                      & 0.189                              & 0.284      \\ 
\hline
Small\_cheese             & 0.887      & 0.942                      & 0.807                              & 0.632      \\
\hline
\end{tabular}
\label{tab:table5}
\end{table}

\subsection{Experiment setting and evaluation}
The computer utilized for our implementation featured a single AMD Ryzen 5 PRO 4650U CPU with 16.0 GB of memory. Our deep learning model employed a multi-layer feed-forward artificial neural network trained using stochastic gradient descent with back-propagation. The learning rate was set to 0.03, and we utilized the Adam optimizer. In order to prevent overfitting, we employed early stopping with a strategy based on score improvement, and a patience of 3. Additionally, we applied regularization with a value of 1 for both L1 and L2. The model consisted of two fully connected layers with ReLU activation functions and 50 neurons, followed by an output layer with Softmax activation and two neurons. To assess the performance of our approach, we employed accuracy, recall, and precision as evaluation metrics. Aside from the model trained merely on the training dataset, we also augment the training dataset using SMOTE Upsampling and autoencoder methods, equalizing the classes. Regarding the autoencoder used to generate synthetical data, augmenting the training dataset, we developed a multi-layer feed-forward artificial neural network trained using stochastic gradient descent with back-propagation. The learning rate, optimizer, and regularization values were set to 0.03, Adam, and value of 1 for both L1 and L2, respectively. Similar to the trained deep neural network, we employed early stopping with a strategy based on score improvement. The encoder employs ReLU activation functions and contains three fully connected layers with eight, four, and two neurons, respectively. The decoder includes ReLU activation functions and contains two fully connected layers with four and eight neurons, along with an output layer (using mean square error loss function with 10 neurons). For evaluation purposes, we implemented 10-fold cross-validation and tested the model on the test dataset. 

\subsection{Performance of models in terms of generalizability}
Table~\ref{tab:table6} provides a summary of the performances of the models using different metrics. As it is apparent, the NSAI model outperforms all other models with the best generalizability, while Deep NN-SMOTE seems to have the lowest performance among the models.

\begin{table}[!htb]
\caption{Performance of models on test data using various sources of learning}
\centering
\small
\begin{tabular}{|l|p{3cm}|l|l|l|l|l|} 
\hline
Models              & Source of learning                    & Accuracy (\%) & \multicolumn{2}{c|}{Recall (\%)} & \multicolumn{2}{c|}{Precision (\%)} \\ 
\cline{3-7}
                    &                                       &               & High        & Low   & High           & Low    \\ 
\hline
Deep NN             & Training data                         & 83.53         & 85.81       & 75    & 93.65          & 54.55  \\ \hline
Deep NN-SMOTE       & Training data + Synthetical data      & 82.35         & 85.51       & 68.75 & 92.19          & 52.38  \\ \hline
Deep NN-Autoencoder & Training data + Synthetical data      & 83.53         & 86          & 68.75 & 92.31          & 55     \\ \hline
NSAI                & Training data + Educational knowledge & 84.71         & 86          & 81    & 95             & 57     \\
\hline
\end{tabular}
\label{tab:table6}
\end{table}

Specifically, the NSAI model could almost achieve an accuracy of 85\%, and recall of 86\% and 81\% for the \textit{high} and \textit{low} performers on the unseen data. This highlights that not only does the NSAI model have the highest likelihood of identifying a significant proportion of \textit{high} performers accurately, but also it is the best model in identifying \textit{low}-performer learners compared to other models. Similarly, the NSAI model has also been shown to have the best performance regarding the precision of \textit{high}-performer learners (meaning out of all learners predicted as \textit{high} performers, 95\% were correct), and it is ranked the best concerning the precision of \textit{low} performers. Because correctly identifying \textit{low}-performing learners is of the utmost importance in our case (and in education given the high-risk nature of the education domain), it is fair to say that a higher number of false positives in the models could be considered acceptable. In other words, the NSAI model's high recall rates suggest that it has the potential to effectively predict \textit{low}-performing learners, which is a valuable characteristic in the field of education (as incorrectly identifying \textit{low} performers could have negative consequences for their future educational outcomes and overall development).

The second-best model in terms of generalizability is the Deep NN model which is trained on original training data. This model exhibits a high ability to correctly predict \textit{high} performers (recall of nearly 86\%) out of all actual samples. However, when it comes to identifying \textit{low} performers, its performance is relatively weaker, classifying only 75\% of them out of all actual samples. On the other hand, the Deep NN model trained on original training data augmented by SMOTE exhibits the poorest performance in terms of generalizability. In practical scenarios, this model is unable to accurately identify \textit{low}-performing learners, achieving a classification rate of less than 70\%. Interestingly, all the deep NN models that were trained on training data (original or both original and synthetical) have appeared to face performance drop when it comes to their generalizability power. However, when evaluating the NSAI model's performance on unseen data, it becomes evident that it has better generalizability due to its learning from both training data and explicit knowledge. The incorporation of explicit knowledge in the NSAI model provides it with a deeper understanding of the underlying relationships among training examples. This enables the model to have more effective learning even in situations where the training dataset is unrepresentative or when the test set exhibits a different distribution compared to the training set. By leveraging explicit knowledge, the NSAI model gains an advantage in its ability to generalize beyond the specific characteristics of the training data. This is particularly beneficial when faced with new or unseen data, as it can draw upon its broader understanding of the domain and the educational context. Finally, regarding the effectiveness of augmenting the training data, while autoencoder augmented data appears to slightly improve some aspects of the model performance in terms of generalizability (e.g., recall of \textit{High} performers), it causes decrement in other aspects like recall of \textit{Low} performers. Interestingly, the SMOTE method even slightly resulted in a performance drop, compared to the original training. 
\subsection{Performance of models in terms of handling data biases and interpretability of predictions}
To determine if trained models learned biases from the training data and incorporated them into their predictions, we employ a two-step approach. Firstly, we conduct a correlation analysis to assess the presence of strong positive or negative correlations between the features and the class label (investigating potential spurious correlations). While this analysis alone cannot definitively establish whether the models genuinely learned to rely on features that are not causally related to the target variable but are strongly correlated with it in the training data or disregarded the existing relationships and patterns among other features during decision-making, examining the internal workings of the models can complement the analysis and provide insights into whether biases were learned and reflected. If the prediction explanation highlights the spurious correlations as the most influential reasons for the models' decision-making process, and we discover that this is the reason for the models' misprediction of the test data, we can conclude that the models failed to grasp certain underlying patterns and relationships due to the presence of biases (represented by the spurious correlations). 

The results of the correlation analysis of both training and test data is shown in Table 4. Obviously, while most features are positively correlated to the label, Hitting\_wall has a negative correlation with the label, meaning the fewer the players bump into the walls, the higher the final performance. Amongst the positive correlations, in the training data, there is a strong positive correlation between consuming Small\_cheese with the Final\_score. Specifically, the Small\_cheese feature has a positive correlation of 0.887 with the class label. This can cause the model learning biases based on such spurious correlations and disregarding the importance of other features and their relationships to the label. As mentioned previously, while the inclusion of the Small\_cheese feature could offer some useful information, there is no causal relationship between this feature and the class label and higher small cheese consumption does not necessarily indicate better CT concepts and skills. When it comes to the test data, while the direction of the correlation is similar, the strengths are weaker compared to the training data. Consequently, the models may heavily rely on features with strong spurious correlations during training, limiting their generalizability to test data that lacks such strong correlations in its distribution. 

To further investigate this matter and provide interpretability over the predictions, we implemented the LIME method on the first three models and extracted rules from the NSAI model. Figure~\ref{fig:myfig5} presents LIME explanations for the model predictions. As can be seen from Figure 5a, the Deep NN model has learned to heavily rely on the feature that is not causally related to the target variable but is strongly correlated with it (i.e., the small cheese). Moreover, it gives a higher weight to features like Arrow and Hitting wall along with Big cheese consumption that are not explicit predictors/indicators of CT knowledge. More importantly, the model has fully ignored considering Loop and Function features and paid little attention to Conditional and Simulation which are all causally related to CT skills and concepts. Similarly, both the Deep NN-SMOTE and Deep NN-Autoencoder have used Small\_cheese followed by Arrow as their primary features in their decision-making, fully ruled out the important feature of Loop, and gave a little attention to the crucial feature of Function and Conditional in their decision making. Consequently, while the models have employed a combination of features in their predictions, the feature with the highest correlations appears to be the primary features for decision making and the model fails to consider some essential features like Loop, Function, and Conditional during the testing stage. 
\begin{figure}
  \centering
  \begin{tabular}{@{}c@{}}
    \includegraphics[width=0.8\textwidth]{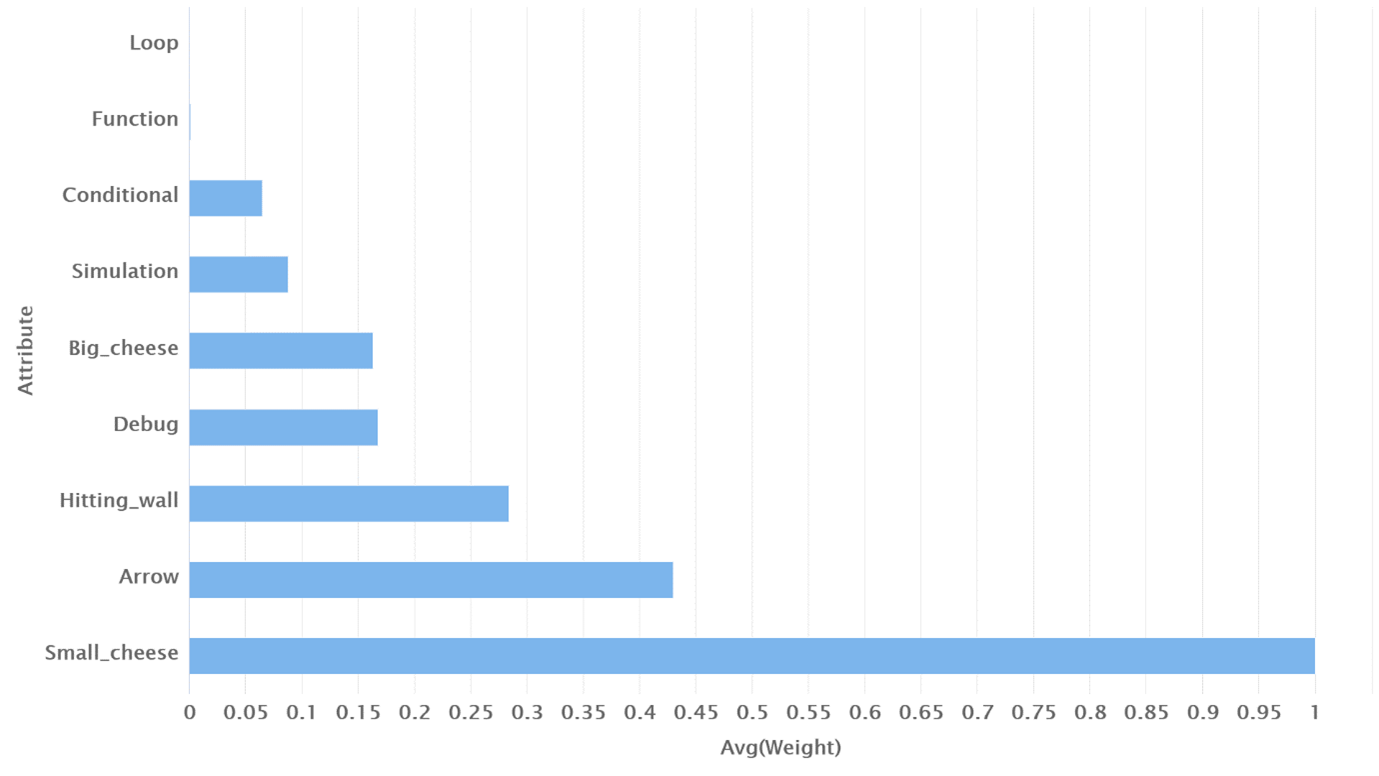} \\
    \normalsize (a)
  \end{tabular}
  \vspace{\floatsep}
  \begin{tabular}{@{}c@{}}
    \includegraphics[width=0.8\textwidth]{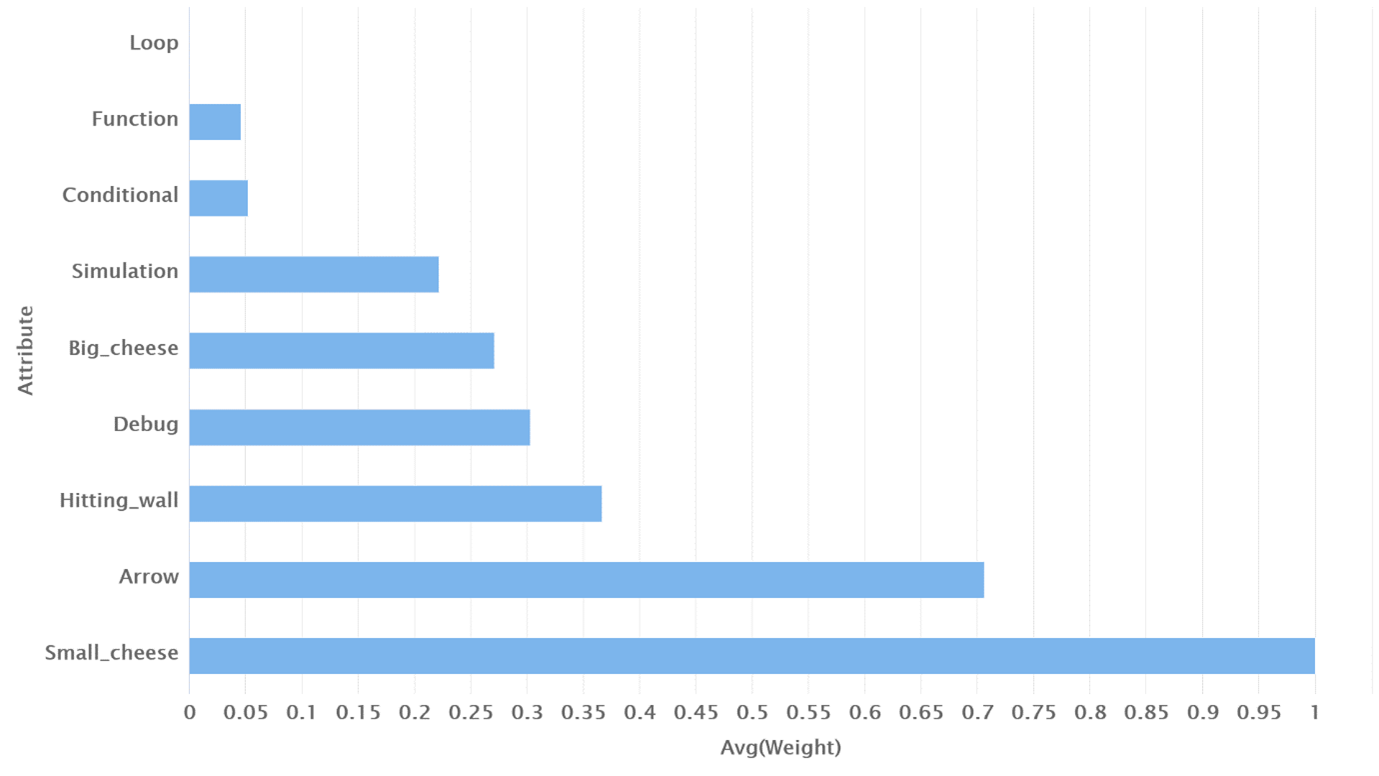} \\
    \normalsize (b)
  \end{tabular}
\end{figure}
\clearpage
\begin{figure}
  \centering
  \begin{tabular}{@{}c@{}}
    \includegraphics[width=0.8\textwidth]{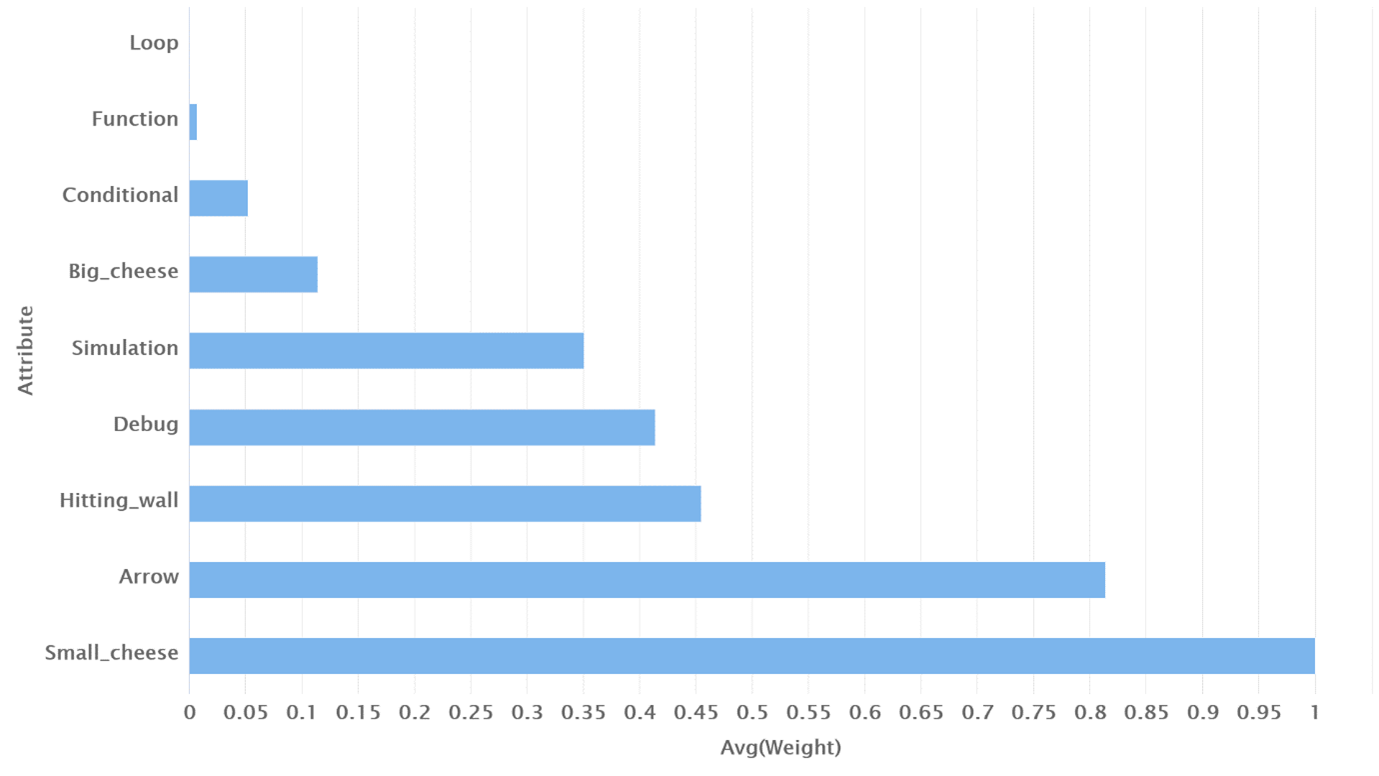} \\[\abovecaptionskip]
    \normalsize (c)
  \end{tabular}
  \caption{Global LIME explanations, a) Deep NN, b) Deep NN-SMOTE, and c) Deep NN-Autoencoder.}
        \label{fig:myfig5}
\end{figure}

Table~\ref{tab:table7} lists three examples in which the models mispredicted the examples during the testing. Obviously, the models incorrectly predicted a final score of \textit{low} as \textit{high} with full confidence due to the high number of cheese consumption (learning the spurious correlations). For instance, the first row of the table shows that the Deep NN model incorrectly predicted a \textit{low}-performer learner as \textit{high} mainly because of their high usage of small cheese and arrow that are not causal predictors of CT knowledge (allocating almost two-thirds of the entire feature importance to these two). Considering the results of the correlation analysis and the global and local LIME explanations, it could be concluded that the models have heavily relied on the learned spurious correlations during training and ignored important factors, limiting their generalizability to test data. In other words, the models learned and reflected biases in their decision-making by learning spurious correlations that caused overlooking the essential/underlying information.

\begin{table}[!htb]
\centering
\caption{Local LIME explanations for some examples of mispredicted cases in the test dataset}
\arrayrulecolor[rgb]{0.494,0.494,0.494}
\small
\begin{tabular}{|p{1.5cm}|p{1.5cm}|p{1.5cm}|p{5cm}|p{4.8cm}|} 

\hline
Model  & \begin{tabular}[c]{@{}l@{}}True value, \\Prediction\end{tabular} & \begin{tabular}[c]{@{}l@{}}Confidence \\(Low, High) \\~\end{tabular} & Supporting                                                                                                          & Contradicting                                                                                                                \\ 
\hline
\multirow{3}{*}{Deep NN} & \begin{tabular}[c]{@{}l@{}}Low,\\High\end{tabular} & \begin{tabular}[c]{@{}l@{}}0.000,\\1.000\end{tabular} & \begin{tabular}[c]{@{}l@{}}Small\_cheese = (Val*=70, Imp=0.465) \\Arrow = (Val=76, Imp*=0.204)\end{tabular} & \begin{tabular}[c]{@{}l@{}}Hitting\_wall = (Val=9, Imp=-0.581) \\Conditional = (Val=0, Imp=-0.162)\end{tabular} \\
\cline{2-5}
& \begin{tabular}[c]{@{}l@{}}Low,\\High\end{tabular} & \begin{tabular}[c]{@{}l@{}}0.000,\\1.000\end{tabular} & \begin{tabular}[c]{@{}l@{}}Small\_cheese = (Val=61, Imp=0.463) \\Arrow = (Val=91, Imp=0.204)\end{tabular} & \begin{tabular}[c]{@{}l@{}}Hitting\_wall = (Val=2, Imp=-0.582) \\Conditional = (Val=2, Imp=-0.162)\end{tabular} \\
\cline{2-5}
& \begin{tabular}[c]{@{}l@{}}Low,\\High\end{tabular} & \begin{tabular}[c]{@{}l@{}}0.000,\\1.000\end{tabular} & \begin{tabular}[c]{@{}l@{}}Small\_cheese = (Val=67, Imp=0.463) \\Arrow = (Val=100, Imp=0.203)\end{tabular} & \begin{tabular}[c]{@{}l@{}}Hitting\_wall = (Val=4, Imp=-0.582) \\Conditional = (Val=0, Imp=-0.163) \\~\end{tabular} \\
\hline
\multirow{3}{*}{\begin{tabular}[c]{@{}l@{}}Deep NN- \\SMOTE\end{tabular}} & \begin{tabular}[c]{@{}l@{}}Low,\\High\end{tabular} & \begin{tabular}[c]{@{}l@{}}0.000,\\1.000\end{tabular} & \begin{tabular}[c]{@{}l@{}}Small\_cheese = (Val=70, Imp=0.397) \\Arrow = (Val=76, Imp=0.282)\end{tabular} & \begin{tabular}[c]{@{}l@{}}Hitting\_wall = (Val=9, Imp=-0.643) \\Conditional = (Val=0, Imp=-0.109)\end{tabular} \\
\cline{2-5}
& \begin{tabular}[c]{@{}l@{}}Low,\\High\end{tabular} & \begin{tabular}[c]{@{}l@{}}0.000,\\1.000\end{tabular} & \begin{tabular}[c]{@{}l@{}}Small\_cheese = (Val=61, Imp=0.397) \\Arrow = (Val=91, Imp=0.282)\end{tabular} & \begin{tabular}[c]{@{}l@{}}Hitting\_wall = (Val=2, Imp=-0.643) \\Conditional = (Val=2, Imp=-0.109)\end{tabular} \\
\cline{2-5}
& \begin{tabular}[c]{@{}l@{}}Low,\\High\end{tabular} & \begin{tabular}[c]{@{}l@{}}0.000,\\1.000\end{tabular} & \begin{tabular}[c]{@{}l@{}}Small\_cheese = (Val=67, Imp=0.397) \\Arrow = (Val=100, Imp=0.282)\end{tabular} & \begin{tabular}[c]{@{}l@{}}Hitting\_wall = (Val=4, Imp=-0.643) \\Conditional = (Val=0, Imp=-0.109)\end{tabular} \\
\hline
\multirow{3}{*}{\begin{tabular}[c]{@{}l@{}}Deep NN- \\Autoencoder\end{tabular}} & \begin{tabular}[c]{@{}l@{}}Low,\\High\end{tabular} & \begin{tabular}[c]{@{}l@{}}0.000,\\1.000\end{tabular} & \begin{tabular}[c]{@{}l@{}}Small\_cheese = (Val=61, Imp=0.339) \\Arrow = (Val=91, Imp=0.290)\end{tabular} & \begin{tabular}[c]{@{}l@{}}Hitting\_wall = (Val=2, Imp=-0.683) \\Conditional = (Val=2, Imp=-0.094)\end{tabular} \\
\cline{2-5}
& \begin{tabular}[c]{@{}l@{}}Low,\\High\end{tabular} & \begin{tabular}[c]{@{}l@{}}0.000,\\1.000\end{tabular} & \begin{tabular}[c]{@{}l@{}}Small\_cheese = (Val=67, Imp=0.339) \\Arrow = (Val=100, Imp=0.290)\end{tabular} & \begin{tabular}[c]{@{}l@{}}Hitting\_wall = (Val=4, Imp=-0.683) \\Conditional = (Val=0, Imp=-0.094)\end{tabular} \\
\cline{2-5}
& \begin{tabular}[c]{@{}l@{}}Low,\\High\end{tabular} & \begin{tabular}[c]{@{}l@{}}0.000,\\1.000\end{tabular} & \begin{tabular}[c]{@{}l@{}}Small\_cheese = (Val=59, Imp=0.339) \\Arrow = (Val=86, Imp=0.290)\end{tabular} & \begin{tabular}[c]{@{}l@{}}Hitting\_wall = (Val=0, Imp=-0.683) \\Conditional = (Val=0, Imp=-0.094)\end{tabular} \\
\hline

\multicolumn{5}{l}{\footnotesize{*Val=Value}}\\
\multicolumn{5}{l}{\footnotesize{*Imp=Importance}}
\end{tabular}
\label{tab:table7}
\end{table}

The extracted rules from the NSAI approach are shown in Table~\ref{tab:table8}. As can be seen, unlike the model explanation provided by the LIME model, not only does the NSAI approach provide a combination of (observable and unobservable) features that contributed to the model’s final decision-making, but also highlights the learned representations (i.e., latent features) and features contributing to their estimations/construction. More explicitly, as the first row of the Table~\ref{tab:table8} shows, learners’ final performance/score in CT is predicted \textit{high} provided that the value of the combination of learned representations of CT concepts and skills as well as heads \footnote{In the NSAI, heads refer to the individual neural network nodes or units within a layer. Specifically, each head represents a single neuron in a layer of the network.} is larger than the threshold of 4.64. In other words, learners’ final performance of CT is predicted \textit{high} if they are good at CT skills and concepts, as well as other heads that are related to learners’ performance with regard to other skills. Given that CT skills (with a weight of 2.00) and concepts (with a weight of 0.83) are both taken into account in the decision-making along with a combination of supporting (e.g., small and big cheeses, hitting wall) and contradictory features (e.g., conditional, loop, arrow, debug, simulation) indicates that the model properly considers skills and concept of CT for its decision making and complies with the introduced causal relationships. Furthermore, in line with the injected knowledge to the model structure during the training, the learned representations of CT concepts mainly consider in their estimations features of conditional and loop (with the weight of 3.00), whereas CT skills mainly take into account features of debug, function and simulation (see the second and third row of tabel). Finally, head 1, 2, and 3 is constructed according to a combination of supporting (e.g., small and big cheeses, hitting wall, simulation) and contradictory features (e.g., conditional, loop, arrow, debug). Consequently, the NSAI model clearly takes into account the given causal relationship related to CT skills and concepts (the educational knowledge presented in the form of rules) and accordingly learns the underlying patterns from the training data. This implies that, unlike the other three models that mostly focus on small cheese consumption and learn spurious correlations to improve the accuracy of the model on test data, the NSAI model took into account educational knowledge and adhered to educational restrictions while learning from the data to improve accuracy on test data. It is worth noting that the extracted rules also allow revising the initial educational knowledge using the training data. In that, CT skills play a more important role in predicting the final performance of learners compared to CT concepts (see the weights associated with CT skills and concepts in the first row of the table). 

\begin{table}
\centering
\caption{Model explanation for the NSAI using rule extraction from neural networks}
\arrayrulecolor{black}
\small
\begin{tabular}{|l!{\color{black}\vrule}p{0.8\linewidth}|} 
\arrayrulecolor{black}\hline
Variables                                                 & Rules                                                                                                                                                                                                                                                  \\ 
\arrayrulecolor{black}\hline
\begin{tabular}[c]{@{}l@{}}Final\_score: \\~\end{tabular} & 4.6377187  2.4166102 * (head2,
  head3) + 0.8252018 * (CT\_concepts) + 2.0046637 * (CT\_skills) + 1.7674259 *
  (head1)                                                                                                                                \\ 
\arrayrulecolor{black}\hline
CT\_concepts:                                             & 4.6082096  0.2453651 *
  (Small\_cheese) + 3.002934 * (Conditional,Loop) + 0.0472862 *
  (Debug,Simulation,Function,Big\_cheese,Hitting\_wall) + -0.07132121 * (Arrow)                                                                                 \\ 
\arrayrulecolor{black}\hline
CT\_skills:                                               & 8.519821  0.20699154 *
  (Small\_cheese) + 2.3430111 * (Simulation) + 1.0791004 * (Function) +
  -0.18917799 * (Conditional,Loop) + 2.6324146 * (Debug) + 0.45198494 *
  (Big\_cheese) + -0.0066499244 * (Arrow) + -0.11537525 * (Hitting\_wall)       \\ 
\arrayrulecolor{black}\hline
head1:                                                    & 2.2751489  -0.070589505 *
  (Conditional,Loop,Debug,Arrow) + 0.80795884 * nt(Big\_cheese) + 0.2296475 *
  (Hitting\_wall) + -0.43813425 * (Function) + 0.09194418 * (Small\_cheese) +
  0.0072831404 * (Simulation)                                    \\ 
\arrayrulecolor{black}\hline
head2:                                                    & 2.881811  -0.43790448 *
  (Function) + -0.04586086 *
  (Conditional,Loop,Debug,Simulation,Arrow,Hitting\_wall) + 0.8505517 *
  (Big\_cheese) + 0.097365424 * (Small\_cheese)                                                                           \\ 
\arrayrulecolor{black}\hline
head3:                                                    & 2.874901  -0.017702527 *
  (Simulation,Hitting\_wall) + 0.8470087 * (Big\_cheese) + -0.4385394 *
  (Function) + 0.09731795 * (Small\_cheese) + -0.06676157 * (Conditional) +
  -0.09061724 * (Loop) + -0.051380966 * (Debug) + -0.031886093 * (Arrow)  \\
\arrayrulecolor{black}\hline
\end{tabular}
\label{tab:table8}
\end{table}

\section{Discussion and conclusion}
This study presents a neural-symbolic AI (NSAI) approach for modelling learners’ computational thinking knowledge that learns from both symbolic educational knowledge and training data. It then compares the performance of the NSAI with deep neural networks trained merely on training data, training data augmented with SMOTE Upsampling, and autoencoder. 

Regarding the generalizability of the models, our findings indicate that the NSAI model, followed by the Deep NN and Deep-Autoencoder, exhibits the best generalizability among the models considered. The NSAI model demonstrates high accuracy and recall on unseen data, suggesting its potential to effectively predict both \textit{high} and \textit{low}-performing learners. In the field of education, correctly identifying \textit{low}-performing learners is crucial for providing timely interventions and support \cite{vincent2020trustworthy}. The high recall rates of the NSAI model make it a valuable tool in this regard, as it minimizes the risk of falsely identifying \textit{low} performers. On the other hand, the other models show poor generalizability, particularly in identifying \textit{low} performers. This indicates the limitations of using traditional deep learning models without additional techniques or knowledge incorporation to address the challenges posed by education data. This finding confirms the argument put forward by Venugopal et al. \cite{venugopal2021neuro} and Hooshyar and Yang \cite{hooshyar2021neural} in that augmenting ANNs with symbolic knowledge can regularize them, improving their generalizability by achieving higher accuracy and scalability by enabling them to learn from smaller datasets. 

The superior generalizability of the NSAI model can be attributed to its incorporation of explicit knowledge in addition to the training data. By leveraging domain-specific knowledge, the NSAI model gains a deeper understanding of the underlying causal relationships among training examples. This broader understanding enables the model to generalize beyond the specific characteristics of the training data, making it more robust when faced with new or unseen data (e.g., Garcez et al., 2022). In educational contexts, where datasets may be unrepresentative or exhibit different distributions, the NSAI model's ability to draw upon its explicit knowledge becomes particularly advantageous. For instance, a study conducted by Shakya et al. \cite{shakya2021student} proposes an innovative approach to enhance automated instruction systems by accurately predicting student strategies and providing personalized support. Their neural-symbolic approach combines symbolic knowledge, using Markov models to represent inherent relationships among input variables, with deep neural networks (specifically LSTMs). The integration of symbolic knowledge enables the model to capture complex relationships and patterns that may not be evident from the training data alone, resulting in improved accuracy and generalizability. These findings align with our research, highlighting the significance of incorporating domain knowledge in machine learning models, particularly neural networks, for educational applications. Regarding the impact of different data augmentation methods, while the autoencoder technique may slightly improve certain aspects of model performance on test data, the SMOTE method can actually cause a minor drop in performance compared to the original training data. This discrepancy can be attributed to a mismatch in the distribution between the training and test data. Data augmentation methods like SMOTE and autoencoder introduce synthetic or reconstructed samples to the training data. If the distribution of the augmented training data deviates from the distribution of the test data, the model may struggle to generalize well. Deep learning models are particularly sensitive to the distribution of the training data, and if training data contains characteristics or patterns that are absent in the test data, the model's performance may suffer significantly. These findings align with the research of Ramezankhani et al. \cite{ramezankhani2016impact}, which suggests that certain data augmentation methods can occasionally result in reduced model performance. Conversely, the NSAI approach appears to outperform all the trained models. 

The analysis of data biases and interpretability provides insights into how the trained models rely on spurious correlations and ignore some essential features during decision-making. The correlation analysis reveals that certain features, such as Small\_cheese, exhibit strong positive correlations with the class label (Final\_score). Models that heavily rely on these strongly correlated features (especially the spurious correlations) during training may overlook the importance of other features and their relationships to the label. As underlined by Zhou et al. \cite{zhou2021examining} and Hutt et al. \cite{hutt2019evaluating}, learning such biases can hamper the generalizability of test data as it causes the model to overlook crucial factors. As our findings show, the deep learning models that were merely trained on data not only mainly rely on the learned spurious correlations during their decision-making, but also mostly rule out many important features that are causally related to learner performance of CT. More explicitly, the models learn data-related biases during training and reflect them in their decision-making. To address these biases and enhance interpretability, the NSAI model incorporates educational knowledge and adheres to educational restrictions during training. The model's learned representations and feature contributions reflect its adherence to the underlying causal relationships between CT skills, concepts, and performance. The extracted rules from the NSAI model provide explicit guidelines for predicting high or low performance based on a combination of observable and unobservable features. These rules not only highlight the importance of CT skills and concepts but also shed light on the relevance of hidden heads that are related to learners' performance in other skills. In other words, the extracted rules from the NSAI approach allow refining our initial educational knowledge as it provides weights for the rules indicating that CT skills are more important in predicting learner performance compared to CT concepts. 

The incorporation of educational knowledge into the NSAI model serves two key purposes. Firstly, it helps alleviate the issue of data biases by providing a framework that ensures the model considers a diverse range of factors beyond spurious correlated features. One of the main objectives of machine learning is to create reliable representations that accurately capture the causal relationship between input features and output labels. However, when models are trained on biased datasets, they may end up paying more attention to spurious correlations between input/output pairs that are not fundamentally relevant to the problem being solved \cite{zhou2021examining}. The NSAI approach not only develops a more comprehensive understanding of learner performance by considering relevant factors, but also addresses the class imbalance issue by learning the underlying relationships and patterns in the training data which ensures a fair representation of all classes and improves the model's overall performance. Secondly, the integration of educational knowledge into the model architecture enables the model to provide interpretable predictions. The model's decision-making process becomes transparent, as it can explain the learned representations that comply with the injected educational knowledge. As highlighted by several recent research (e.g., \cite{conati2018ai,hooshyar2021neural,meltzer2022european,vincent2020trustworthy}), such interpretability is crucial in educational contexts, where stakeholders require explanations for model predictions to ensure transparency, fairness, and trustworthiness. 

By combining the incorporation of knowledge with deep learning techniques, the NSAI model demonstrates how interpretability and generalizability can be improved in educational machine learning applications. The model's ability to capture complex relationships, consider multiple factors, and provide rule-based explanations facilitates a deeper understanding of learner performance. This not only aids in accurate predictions but also assists educators in identifying specific areas for intervention and support \cite{conati2018ai,fiok2022explainable}. In conclusion, The NSAI model emerges as a promising approach, showcasing better generalizability and the ability to leverage domain-specific knowledge for trustworthy and interpretable predictions in educational contexts. Further research and development in this direction can contribute to the advancement of machine learning techniques in education, enabling more effective support for learners and educators.

\subsection{Limitations and future works}
One limitation of this work is the lack of experimentation using different datasets and exploring various deep neural network variants. Future research can address this by testing the proposed NSAI approach on different educational tasks, such as sequential and temporal analysis, to enable early prediction of learner performance and identification of at-risk learners. Additionally, future work can focus on experimenting with different ways to inject educational knowledge into neural networks, potentially leveraging neural symbolic AI methods. Finally, it is desirable to integrate the NSAI-based learner modelling method into existing digital learning platforms and evaluate its effectiveness in real-world classrooms, providing personalized learning that is impartial, trustworthy, and interpretable.

\bibliographystyle{unsrt}  

\end{document}